\documentclass[10pt,twocolumn,letterpaper]{article}

\usepackage[pagenumbers]{cvpr} 
\usepackage{graphicx}
\usepackage{amsmath}
\usepackage{amssymb}
\usepackage{booktabs}
\usepackage{multirow}
\usepackage{balance}
\usepackage{float}
\usepackage{amsthm}
\usepackage{amsmath}
\usepackage{algorithm}
\usepackage{algorithmic}

\usepackage[pagebackref,breaklinks,colorlinks]{hyperref}

\usepackage[capitalize]{cleveref}
\crefname{section}{Sec.}{Secs.}
\Crefname{section}{Section}{Sections}
\Crefname{table}{Table}{Tables}
\crefname{table}{Tab.}{Tabs.}

\def\confName{CVPR}
\def\confYear{2024} 
\begin{document} 
\title{Causal Perception Inspired Representation Learning for Trustworthy Image Quality Assessment}
\author{Lei Wang, Desen Yuan\\
University of Electronic Science and Technology of China\\
{\tt\small wangleiuestc@outlook.com, desenyuan@gmail.com}
}
\maketitle
\begin{abstract}
Despite great success in modeling visual perception, deep neural network based image quality assessment (IQA) still remains unreliable in real-world applications due to its vulnerability to adversarial perturbations and the inexplicit black-box structure. In this paper, we propose to build a trustworthy IQA model via Causal Perception inspired Representation Learning (CPRL), and a score reflection attack method for IQA model. More specifically, we assume that each image is composed of Causal Perception Representation (CPR) and non-causal perception representation (N-CPR). CPR serves as the causation of the subjective quality label, which is invariant to the imperceptible adversarial perturbations. Inversely, N-CPR presents spurious associations with the subjective quality label, which may significantly change with the adversarial perturbations. To extract the CPR from each input image, we develop a soft ranking based channel-wise activation function to mediate the causally sufficient (beneficial for high prediction accuracy) and necessary (beneficial for high robustness) deep features, and based on intervention employ minimax game to optimize.  Experiments on four benchmark databases show that the proposed CPRL method outperforms many state-of-the-art adversarial defense methods and provides explicit model interpretation.
\end{abstract}

\section{Introduction}  
 
\begin{figure}[!t]
  \centering 
  \includegraphics[width=0.45\textwidth]{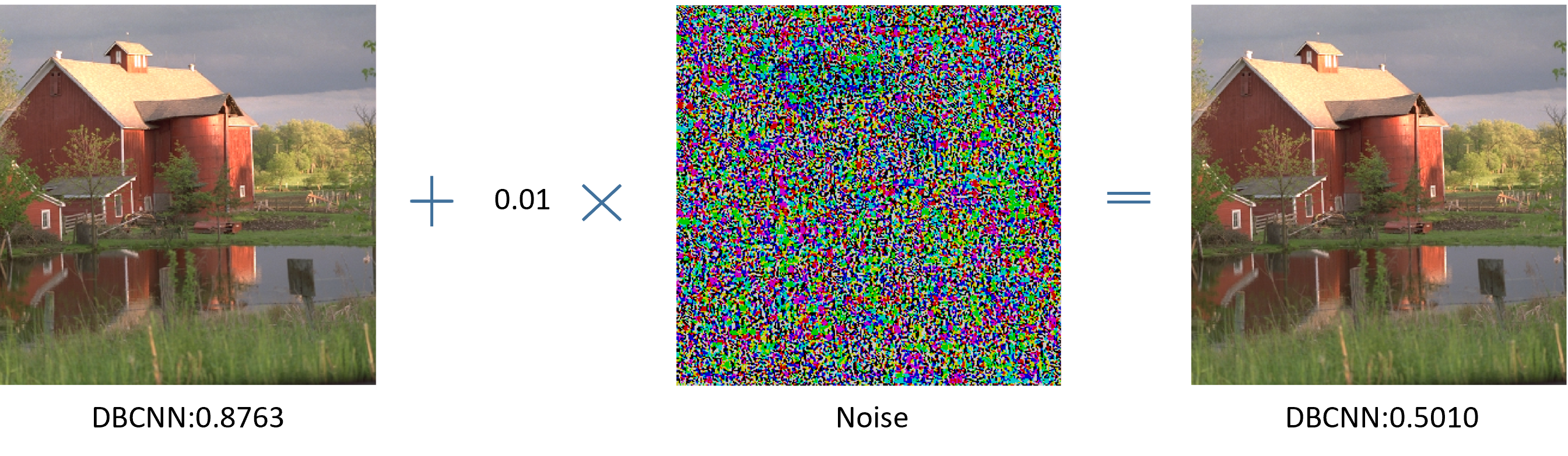} 
  \caption{Demonstration of IQA model adversarial example generation). By adding an imperceptible perturbation, we can drastically change the predicted score of an IQA model for an image.}\label{fig:intro}
\end{figure}

The rapid growth of image data on the Internet has made the automatic evaluation of image quality a vital research and application topic. Objective image quality assessment (IQA) methods can be divided into three categories based on the availability of the original undistorted images: full-reference (FR), reduced-reference (RR), and no-reference/blind (NR/B). Among them, BIQA model is a challenging computational vision task, which mimics the human ability to judge the perceptual quality of a test image without requiring the original image content\cite{wang2004image,min2019quality,zhai2020perceptual,lin2019kadid,ma2021blind}.
However, existing BIQA methods are not reliable, and minor perceptual attacks can fool the quality evaluator to produce incorrect output. This vulnerability affects the security issues, resulting in lower reliability of IQA\cite{zhang2022perceptual}.
As shown in Fig \ref{fig:intro}, after adding minor perturbations, the prediction results of the deep IQA model showed significant errors while the human eye does not perceive the change in quality. 
In image classification, this is known as adversarial attack (e.g., `norm constrained attack). A reliable IQA model should not be overly sensitive to adversarial attacks, that is, be adversarially robust\cite{ilyas2019adversarial,liu2023towards,kaddour2022causal}.

The prevalent approach to adversarial robustness is still empirical adversarial training and its variants. However, adversarial training is a form of “passive immunity” that cannot adapt to dynamic attacks. Model-based denoising defense is another important defense strategy. However, in quality assessment tasks, denoising models may backfire and have adverse effects. This is because the denoising operation itself alters image quality without affecting image semantics\cite{xie2019feature}.
slight interference from the picture will not affect the perceptual quality of the human eye. The specific reason can be related to the all or nothing principle~\cite{kalat2015biological}. Recently, researchers have analysed adversarial examples from non-robust features and robust features, and demonstrated that robust features can still provide precise accuracy even in the presence of adversarial perturbations\cite{ilyas2019adversarial}. Researchers believe that the high-dimensional dot product operation and multi-layer structure of DNN form a dense mixture so that small changes in features can accumulate~\cite{allen2022feature}, and cause excessive Lipschitz constant\cite{zhang2022rethinking}.

We propose a new way to understand the IQA task by using a causal approach. We use a causal framework that shows how the images in the IQA dataset are produced by three factors: reference images, causal perception representation (CPR) and non-causal perception representation (N-CPR). We say that N-CPR are confounders that do not affect image quality but can create false correlations. These confounders can make the model predictions wrong, and they can be any feature that is related to some labels, such as local texture, small edges, and faint shadows. 
\cref{fig:SCM} illustrates the structural causal model (SCM) that we use to describe how the data for the IQA dataset is generated. The normal IQA training and testing can be seen as changing the reference image R while keeping the N-CPR fixed. In this case, the non-robust features are stable, so the model can work well from the training to the test scenario. But in an adversarial scenario, someone can change the image quality by changing the N-CPR without changing how the image looks. This is a problem, because the current IQA networks cannot tell the difference. They do not use any prior knowledge to remove the confounders and stop the false correlations. 

we propose to build a trustworthy IQA model via Causal Perception inspired Representation Learning (CPRL). specifically, IQA models that only depend on statistical associations are inadequate and unreliable, our method uses a causal intervention to break the false correlations and find the true cause of image quality. To extract the CPR from each input image, we develop a soft ranking-based channel-wise activation function to mediate the causally sufficient (beneficial for high prediction accuracy) and necessary (beneficial for high robustness) deep features. We train this module with the probability of necessity and sufficiency (PNS) risk\cite{tian2000probabilities,wang2021desiderata,yang2023invariant}. Finally, we systematically compare the performance of our method with existing methods in different scenarios. Experimental results show that channel activation improves the accuracy of image quality significantly.


\section{Related Work}
\subsection{Image quality assessment}
Image quality assessment aims to obtain an objective model\cite{wang2004image} to predict the image quality, making it close to the subjective quality score. 
Interpretable and reliable structural similarity (SSIM)\cite{wang2004image} and peak signal-to-noise ratio (PSNR) are favored by researchers as a reference image quality method.
However, researchers have been looking for methods similar to the human visual system(HVS) that do not require reference pictures which have been called no-reference(NR) IQA methods\cite{wang2011reduced}. 
Due to the lack of information on reference pictures, naturally, the researchers used the statistical information of pictures as a reference, called natural scene statistics (NSS)\cite{fang2014no}, and proposed Natural Image Quality Evaluator (NIQE) \cite{mittal2012making}. 
Benefiting from the development of deep learning, some learning-based image quality models~\cite{ma2017dipiq, zhang2018blind, bosse2017deep} have obtained stronger performance than traditional quality evaluation methods\cite{min2018blind} due to the powerful feature extraction capabilities of convolution kernels\cite{he2016deep}.
In order to explore more feature expression capabilities, researchers proposed many other methods based on Generative adversarial network~\cite{ma2021blind}, Variational Auto-Encoder~\cite{wang2020blind}, and Transformers~\cite{you2021transformer}.
However, powerful feature representation is difficult to explain, so some researchers try to figure out whether the quality evaluation is decoupled from the content\cite{li2018has}.
Therefore, this paper focuses on the robustness exploration of deep IQA models, and dedicates to discovering efficient ways to improve the robustness of IQA models.

\subsection{Vulnerability of deep neural networks}
The vulnerability of Deep Neural Networks is believed to be due to the differences between the features extracted by the model and the human visual system \cite{ilyas2019adversarial}. So small changes can make deep models ineffective \cite{szegedyIntriguingPropertiesNeural2014, goodfellowExplainingHarnessingAdversarial2015a}.  The most classical anti-sample attack method, Fast Gradient Sign Method (FGSM)\cite{goodfellowExplainingHarnessingAdversarial2015} , which is based on gradient attack, has the following form:
\begin{equation}
x^{\mathrm{adv}}:=x+\varepsilon \cdot \operatorname{sign}\left(\nabla_{x} L\left(h(x), y_{\text {true }}\right)\right)
\end{equation}
where $x^{\mathrm{adv}}$ is the adversarial example, which equals to the original sample plus a small perturbation. $\varepsilon$ The direction of the perturbation is determined by the gradient of the predictor output $h(x)$ and the label $y$ with respect to the loss $L$.  The magnitude of the perturbation is given by $\varepsilon$.
Researchers found that the parameters of the neural network have a dense mixture of terms \cite{allen2022feature}, and adversarial training is doing feature purification which alleviates vulnerability.

\subsection{Adversarial robust models}
A more robust model is not only more consistent with human perceptual properties\cite{tsipras2018robustness,engstrom2019adversarial}, but also increases the ability to generalize outside the domain\cite{shi2020informative}.    
From the point of view of data preprocessing, Feature Squeezing~\cite{xu2017feature} distinguishes between adversarial and clean samples by reducing the color bit depth of each pixel. 
From the point of view of feature extraction, Feature denoising~\cite{xie2019feature} reduces noise in latent features. from the empirical results, it can be seen that the feature spectrum is purer than the original deep network.  
Cisse proposed Parseval networks~\cite{cisse2017parseval}, a layerwise parameters regularization method for reducing the network’s sensitivity to small perturbations by carefully controlling its global Lipschitz constant. Qin proposed a Local Linearization regularizer~\cite{qin2019adversarial} that encourages the loss to behave linearly in the vicinity of the training data. Kannan introduce enhanced defenses using a technique called logit pairing~\cite{kannan2018adversarial}, a method that encourages logits for pairs of clean examples and their adversarial counterparts to be similar. 
All the above methods add additional constraints, such as controlling the Lipschitz constant to suppress the amplification effect of the network, or constraining the local linearity of features, or generating a purpose for feature alignment. 

\subsection{Causality inference}
Various methods have been proposed to achieve causal inference, such as Matching Methods\cite{stuart2010matching}, Propensity Score based Methods\cite{zhang2021deep}, and Reweighting Methods\cite{kuang2020data,shen2018causally}. However, the balancing methods are not applicable in adversarial scenarios, because they mostly rely on the assumptions that the dataset is sufficient while non-perceptual representation are observed in natural images. Moreover, for high-dimensional images, there is not enough data for statistical analysis of causal estimation. Recently, some researchers have leveraged causal inference tools to optimize models from observable high-dimensional data~\cite{yang2023invariant,wang2021desiderata,zhang2023towards}. Correspondingly, many perspectives have emerged to understand adversarial robustness from causal inference~\cite{kaddour2022causal,liu2023towards,kim2023demystifying,tang2021adversarial,zhang2021causaladv}. Furthermore, researchers use causal inference to obtain more reliable representations\cite{lv2022causality,cai2023learning,yang2023specify}.

\begin{figure}
  \centering
  \includegraphics[width=0.4\textwidth]{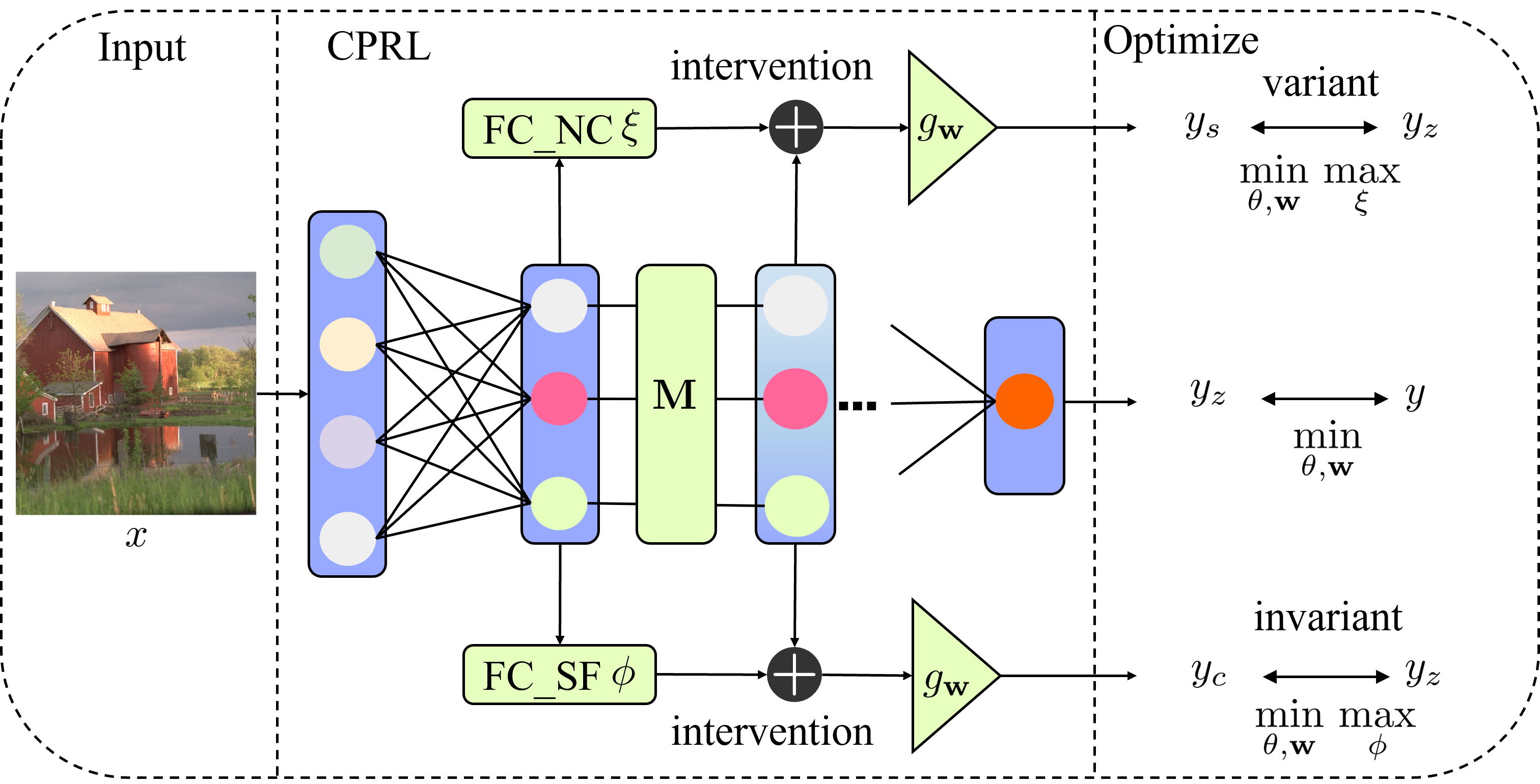}
  \caption{The CPRL module. We simply use this module to replace partial ReLU layers in the ResNet backbone to perform the IQA task.}\label{fig:model}
\end{figure}
\section{Preliminaries}\label{sec:pr}
\subsection{Trustworthy IQA in adversarial attack}

To form a credible IQA task in adversarial scenarios, we assume that there is a dataset $D$ consisting of tuples of images $x_i$, corresponding labels $y_i$, and reference images $R_i$. We can represent the data as $D = \{(R_i,x_i,y_i)^N_{i=1}\}$, where $N$ is the number of samples. The IQA task can be divided into two types: no-reference (NR) and full-reference (FR), depending on whether a reference image is given or not. A scene is FR if the reference image is observable during both training and testing phases. If the reference image is unobservable, then the scene is NR. 
The purpose of the adversarial attack defined in this paper is to find a perturbation $\Delta x$ that reduces the network prediction quality score $f(x)$, which can be expressed as:
\begin{equation}
\underset{\Delta x}{\arg \max } \ L (y,f(x+\Delta x)) \text {, s.t. } d(\Delta x) \leq \epsilon \text {, }
\end{equation}
where $d$ is a signal fidelity measure, $\epsilon$ is the bound of small perturbations, and $L(\cdot,\cdot)$ represents the loss criterion for output and label. In practice, we define $d$ as the $\ell_{\infty}$-norm measure $\| \cdot \|_{\infty}$ and bounded by $\epsilon = 1/255$.
Following the above formula, we clarify the differences between adversarial scenarios and traditional IQA tasks. Let $P(\cdot)$ represent a marginal distribution. The difference between them is that the image $X$ and its marginal distribution $P(X)$ do not satisfy the assumption of independent and identically distributed (i.i.d.) samples, that is, for traditional IQA tasks, $P(X_{\text{train}}) = P(X_{\text{test}})$. For adversarial scenarios, $P(X_{\text{train}}) \neq P(X_{\text{test}})$.

\begin{figure}
      \centering
	   \begin{subfigure}{0.45\linewidth}
		\includegraphics[width=\linewidth]{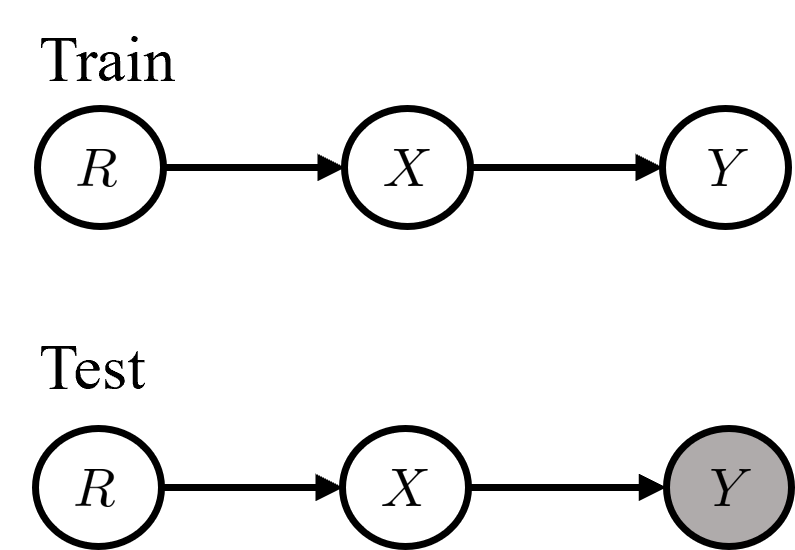}
		\caption{FR-IQA}
		\label{fig:subfig1}
	   \end{subfigure}
	   \begin{subfigure}{0.45\linewidth}
		\includegraphics[width=\linewidth]{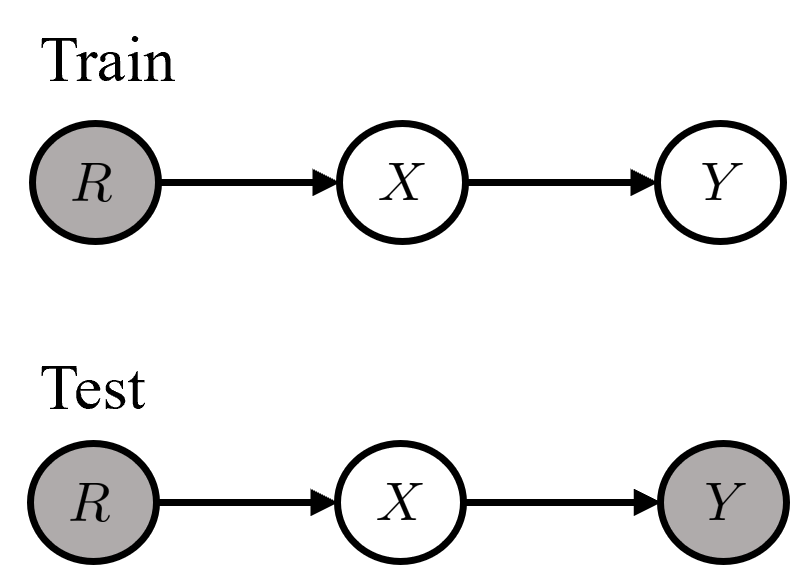}
		\caption{NR-IQA}
		\label{fig:subfig2}
	    \end{subfigure}
	\vfill
	     \begin{subfigure}{0.3\linewidth}
		 \includegraphics[width=\linewidth]{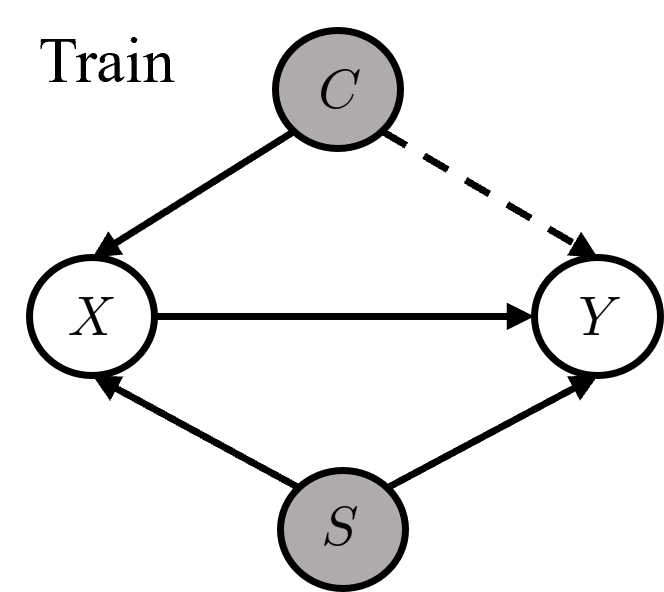}
		 \caption{IQA under spurious correlations}
		 \label{fig:subfig3}
	      \end{subfigure}
	       \begin{subfigure}{0.3\linewidth}
		  \includegraphics[width=\linewidth]{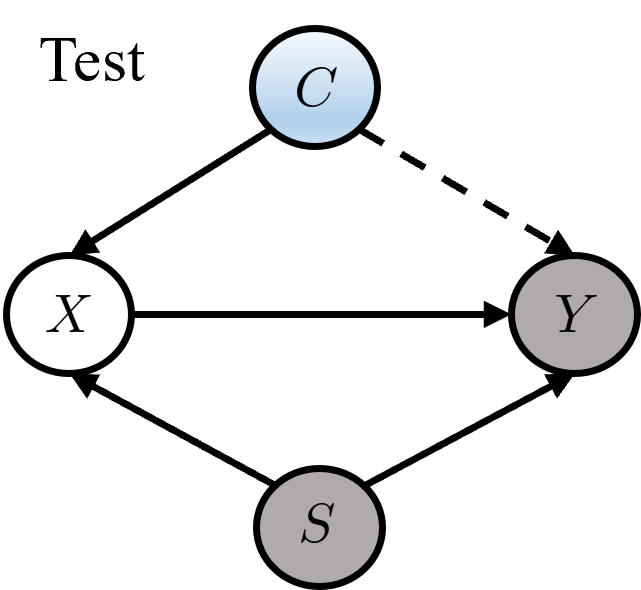}
		  \caption{IQA under adversarial attacks}
		  \label{fig:subfig4}
	       \end{subfigure}
	       \begin{subfigure}{0.3\linewidth}
		  \includegraphics[width=\linewidth]{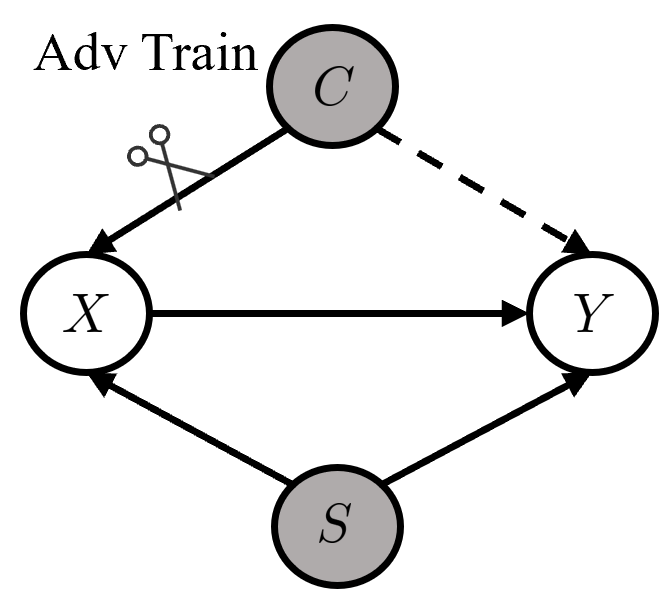}
		  \caption{IQA under adversarial training}
		  \label{fig:subfig5}
	       \end{subfigure}        
	\caption{Causal graphs of IQA learning. Grey nodes represent unobserved variables. (a) Cause graphs of the traditional FR-IQA task during training and testing phases. (b) Cause graphs of the traditional NR-IQA task during training and testing phases. (C) Cause graphs of NR-IQA learning under spurious correlations during the training phase. Due to shortcut learning, the network model learns spurious correlations from the path $X\leftarrow C\rightarrow Y$. (d) Cause graphs of NR-IQA learning during the testing phase. The non-aware variables in blue are not from the same distribution as the training set. During the test phase, the distribution of confounding variables $C$ in blue changes (adversarial attacks). (e) Cause graphs of NR-IQA adversarial learning.  Adversarial training obtains new samples $X$ by changing the confounding variable $C$ and adding them to the training. Think of it as a backdoor intervention.}
	 \label{fig:SCM}
\end{figure}

\subsection{Causal graphs in IQA learning}
A causal graph is a directed acyclic graph in which nodes represent variables and directed edges represent the causal relationship between two variables. We treat images, labels, reference images, causal perception representation (CPR) and non-causal perception representation (N-CPR) as nodes in a causal graph. Our causal graph reveals the underlying statistics of IQA learning, as we will show in the following sections.
As shown in \ref{fig:SCM}, we denote the image as $X$, the mean opinion score (MOS) as $Y$, CPR as $S$, and N-CPR as $C$. The edges are listed as follows: $X \rightarrow Y$ means that MOS $Y$ is derived from image $X$; $R\rightarrow X$ means that the reference image $R$ has a causal influence on image $X$; $C\rightarrow Y$ with dotted line means that the N-CPR variable $C$ has a spurious correlation with MOS $Y$; $C\rightarrow X$ means that the N-CPR variable $C$ has a causal influence on image $X$; $S\rightarrow X$ means that the CPR variable $S$ has a causal influence on image $X$; $S\rightarrow Y$ means that the CPR variable $S$ has a causal influence on MOS $Y$.
For traditional IQA learning in \ref{fig:SCM}(a)(b), its causal graph contains the causal relationship between the image variable $X$, the label $Y$, and the reference image $R$. During the training phase of supervised learning, both the image $X$ and the label $Y$ are observable, and we use them to learn the causal effects in the model. But in the testing phase, only the image $X$ and the causal path are known, and $Y$ is the label to be predicted. For NR-IQA learning, $R$ is unobservable, while for FR-IQA learning, $R$ is observable. For the trustworthy IQA learning scenario, we model an unobserved variable, the non-perceptual variable $C$, and they are illustrated in \ref{fig:SCM}(c). Their differences in the test phase are shown in \ref{fig:SCM}(d). Due to changes in the non-perceptual variable $C$, $X$ may have an indirect effect on $Y$ through the causal path $X\leftarrow C\rightarrow Y$, establishing a spurious correlation with $C$. 
The purpose of trustworthy IQA is to establish true quality-related correlations based on image quality evaluation, independent of non-perceptual variables.

\section{Methodology}\label{sec:mt}

In this section, we present a causal framework to analyze the biases induced by non-perceptual features in IQA, a causal intervention method to eliminate them, and the implementation details and pipeline of our approach.
\subsection{Non-perceptual-feature induced bias in IQA}
According to the causal graph, we can explain why the IQA model fails in adversarial scenarios. We attribute this phenomenon to biases induced by non-perceptual features. In \ref{fig:SCM}(c), the causal path $X\leftarrow C\rightarrow Y$ is a fork structure, and this path is spurious because it is outside the causal path from $X$ to $y$. Opening this path will create non-perceptual feature-induced biases and produce erroneous outcomes.
We give a simple example to illustrate this point. A causal graph A country’s per capita chocolate consumption $\leftarrow$ Economic conditions $\rightarrow$ Number of Nobel Prize winners There is a strong correlation between a country’s per capita chocolate consumption and its number of Nobel Prize winners. This correlation seems absurd because we cannot imagine winning a Nobel Prize for eating chocolate. A more plausible explanation is that more people eat chocolate in wealthy Western countries, and Nobel laureates are preferentially selected from these countries. But this is a causal explanation, which leads to the observed correlation between chocolate and Nobel Prizes. If we could control for the confounding factor of economic conditions and collect data on the poor economic conditions under which chocolate is rarely consumed, then we could come to the correct conclusion that chocolate consumption is not related to the number of Nobel Prize winners.
In \ref{fig:SCM}(e), The fork structural causal path $X\leftarrow C \rightarrow Y$ can be causally intervened through backdoor adjustment to eliminate the bias induced by non-perceptual features in IQA and estimate the causal effect from $X$ to $Y$. We can get the expression of $P(Y | do (X))$ according to the backdoor criterion\cite{pearl2009causality}:
\begin{equation}
P(Y \mid do(X))=\sum_c P(Y \mid X, c) P(C)
\end{equation}
We find that adversarial training can be understood as a backdoor adjustment to a simplified situation:
\begin{equation}
\begin{aligned}
&P(Y \mid do(X))=P(Y \mid X, c_{adv}) P(c_{adv})\\
&+P(Y \mid X, c_{natural}) P(c_{natural})
\end{aligned}
\end{equation}
For a single sample $x$, we convert the summation term into two samples. Natural samples and corresponding adversarial samples are added to the training to eliminate the dependence on $c$. In the loss function of adversarial training, the weighted values of adversarial samples and natural samples can be regarded as the prior probability of the above formula.
However, adversarial training is only a passive defense strategy. We seek more advanced causal representations. The key idea is to utilize the probability of causation (POC) to systematically describe the importance of features.

\subsection{Causal perception inspired representation learning for IQA}
In this subsection, we introduce the score reflection attack method, and a novel channel-wise activation function based on soft ranking and a max-min game training strategy based on PNS, which aims to learn a causal representation that is both sufficient and necessary for the prediction. The overall framework is shown in \ref{fig:model}.

Since image quality results value ranking, when we attack the image quality evaluation task, the aim is to invert the quality of image predictions (relative to the median value of 0.5) to interfere with the ranking, making the score gap larger, we called the score reflection attack:
\begin{equation}
x^{\mathrm{adv}}:=x+\varepsilon \cdot \operatorname{sign}\left(\nabla_{x} \|h(x), \operatorname{sign} (y_{\text {true }}-0.5)\|_{2}\right)
\label{equ_FGSM}
\end{equation}
We first compute a perceptual score $\mathbf{M}$ for the feature $f(x)$ and apply a nonlinear function $g$ to the element-wise product of $f(x)$ and $\mathbf{M}$:  
\begin{equation} 
y=g(f(x) \odot M)
\end{equation}
We use a sigmoid function to obtain $\mathbf{M}$ from the sort index of $f(x)$ with and bias term $b$, and Number of channels $K$:
\begin{equation}
M=\frac{1}{1+e^{-(sort(f(x))-K/2+bK)}} 
\end{equation}
To ensure that the activated channels are both necessary and sufficient for the prediction, we adopt the probability of necessity and sufficiency (PNS) as a measure of causal relevance and optimize it during training\cite{yang2023invariant}. 
PNS is defined by Pearl as the joint probability of a feature being both necessary and sufficient for the outcome:
\begin{equation} 
PNS=P(Z=z, Y=y) \cdot P N+P(Z \neq z, Y \neq y) \cdot P S
\end{equation} 
where $PN$ is the probability of necessity and $PS$ is the probability of sufficiency, given by:
\begin{equation}
\begin{aligned}
& P N  \triangleq P(Y(\boldsymbol{Z} \neq \boldsymbol{z}) \neq y \mid \boldsymbol{Z}=\boldsymbol{z}, Y=y), \\
& P S  \triangleq P(Y(\boldsymbol{Z}=\boldsymbol{z})=y \mid \boldsymbol{Z} \neq \boldsymbol{z}, Y \neq y).
\end{aligned}
\end{equation} 
However, computing counterfactual probabilities is intractable in general. Therefore, we use a lower bound of PNS as our objective, given by\cite{wang2021desiderata}:
\begin{equation}
PNS \geq P(Y=y \mid \operatorname{do}(\boldsymbol{Z}=\boldsymbol{z}))-P(Y=y \mid \operatorname{do}(\boldsymbol{Z} \neq \boldsymbol{z})) 
\end{equation}

We want to maximize PNS during training to make the feature causally relevant. To this end, we introduce two distributions that represent different interventions on the feature $f(x)$. Specifically, the distribution $P^\phi(c \mid \mathbf{X}=\mathbf{x})$ samples a feature $c$ that preserves the perceptual representation of $f(x)$ but changes the non-perceptual representation. The distribution $P^{\xi}(s \mid \mathbf{X}=\mathbf{x})$ samples a feature $s$ that alters the perceptual representation of $f(x)$ but keeps the non-perceptual representation intact. The distribution $P^{\mathbf{w}}(y|z)$ gives the prediction of $y$ given $z$. Then, we define the PNS risk as the sum of the sufficiency risk $SF$ and the necessity risk $NC$:
\begin{equation}
\begin{aligned}
&R(\mathbf{w}, \phi, \xi)=   S F(\mathbf{w}, \phi)+N C(\mathbf{w}, \xi)
\end{aligned}
\end{equation}
The sufficiency risk $SF$ measures the expected prediction error when $z$ is sampled from the distribution that preserves the perceptual representation. The necessity risk $NC$, on the other hand, measures the expected prediction difference when $z$ is sampled from the distribution that changes the perceptual representation. The risk increases when the representation contains less necessary and sufficient information.

We integrate the perceptual score term $\mathbf{M}$ into the PNS risk, which leads to the final form of $R(\mathbf{w}, \phi, \xi)$ as follows:
\begin{equation}
\begin{aligned}
\mathbb{E}_{(\mathbf{x}, y) \sim \mathcal{D}} \mathbb{E}_{\mathbf{c} \sim P^\phi(\mathbf{c} \mid \mathbf{x})} \mathbb{E}_{\mathbf{s} \sim P^\xi(\mathbf{s} \mid \mathbf{x})}  \Vert y_c -y \Vert_2^2 -  \Vert y_s -y \Vert_2^2 
\end{aligned}
\end{equation}
where $y_c=\operatorname{sig}(g_{\mathbf{w}} \circ  h_\phi(f(x),\mathbf{M}))$, $y_s=\operatorname{sig}(g_{\mathbf{w}} \circ h_\xi(f(x),\mathbf{M}))$.
We use logistic regression as the prediction model and use the score term $\mathbf{M}$ to control the intervention scope. Since we add the conventional regression loss at the end, we replace the labels here with the predicted values without intervention. We have $y=\operatorname{sig}(  g_{\mathbf{w}}   (f(x) \odot \mathbf{M}) ) $. The function $h_\phi(f(x),\mathbf{M}))$ modifies the feature map $f(x)$ by applying a non-linear transformation $\operatorname{FC}_{\phi}$ to the part that is masked by $\mathbf{M}$ and adding it back to the original feature map. The function $h_\xi(f(x),\mathbf{M}))$ does the same thing but to the part that is masked by $1-\mathbf{M}$, which is the complement of $\mathbf{M}$. Formally, we have:
\begin{equation}
\begin{aligned}
& c= h_\phi(f(x),\mathbf{M}))=  \operatorname{FC}_{\phi} f(x) \odot (1-\mathbf{M}) +f(x)\odot   \mathbf{M}, \\
& s= h_\xi(f(x),\mathbf{M}))=\operatorname{FC}_{\xi}f(x)  \odot \mathbf{M}+f(x)\odot  \mathbf{M}.
\end{aligned}
\end{equation}
We want to minimize PNS under any intervention distributions $P^{\xi}$ and $P^{\phi}$, so the final loss is in the form of a min-max game. In other words, the intervention distribution at this time is equivalent to the worst-case adversarial distribution. We want to maximize the representativeness of PNS under this estimated adversarial distribution. We combine the regular MSE loss term with the PNS risk term to get the following optimization objective:
\begin{equation}
\begin{aligned}
\underset{\mathbf{\theta},\mathbf{w}}{\min} \ \underset{\phi,\xi}{\max} \frac{1}{N}\sum_{i=0}^N \Vert g(f(x_i) \odot M) -y_i \Vert_2^2+  R (\mathbf{w}, \phi, \xi)
\end{aligned}
\end{equation}
In addition, spectral normalization\cite{miyato2018spectral} be applied to $\phi$ and $\xi$, Prevent parameters from being too large. We summarize the overall CPRL training procedure in ~\ref{al:CPRL}. The actual implementation adopts an alternating optimization scheme for updating the distribution parameters of the network, and leverages an intervention indicator to switch the output mode of the network and the optimization mode of the training pipeline.

\begin{algorithm}[!t]
\caption{CPRL Training Algorithm}
\label{al:CPRL}
\textbf{Input}: Training data $\mathcal{D} = \{(x_i, y_i)\}^n_{i=1}$. \\
\textbf{Output}: model parameter $\{\theta,w,\phi,\xi\}$. intervention symbol $\text{int}=\text{None}$. Number of iterations $t=0$.
\begin{algorithmic}[1]
\STATE Initialize model with parameters $\{\theta, w, \phi, \xi\}$, $int==\text{None}$.
\WHILE{$t < T$}
\STATE  Calculate the model output and update phi $\{\theta,w\}$ through minimize loss.
\IF{$\text{int}==\text{SF}$}
\STATE  Calculate the model output under $c$ and update $\phi$ through maximize loss,
\STATE  Calculate the model output under $c$ and update $\{\theta,w\}$ through minimize loss,
\ENDIF
\IF {$\text{int}==\text{NC}$}
\STATE  Calculate the model output under $S$ and update $\xi$ through maximize loss,
\STATE  Calculate the model output under $S$ and update $\{\theta,w\}$ through minimize loss, 
\ENDIF
\STATE $t=t+1$, and change $\text{int}$ by t.
\ENDWHILE
\end{algorithmic}
\end{algorithm}

\begin{figure}  
	\centering
	\subfloat[\label{fig:simple a}]{
		\includegraphics[width=0.2\textwidth]{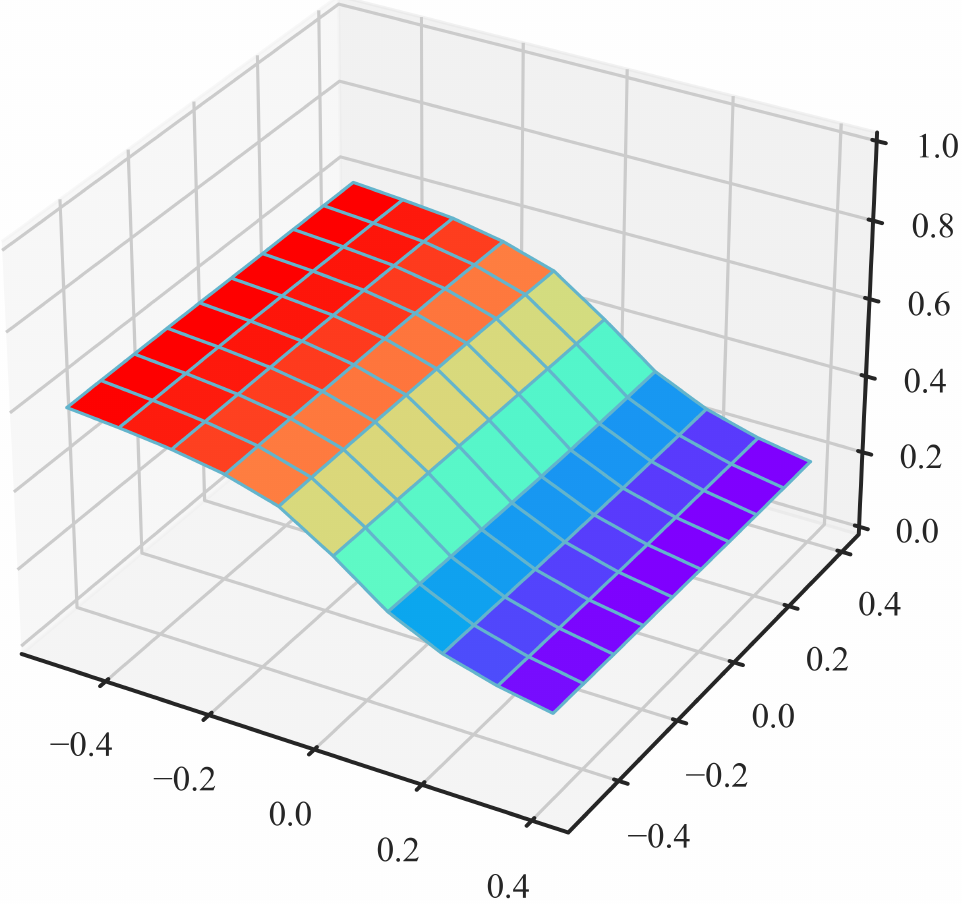}}
	\subfloat[\label{fig:simple b}]{
		\includegraphics[width=0.2\textwidth]{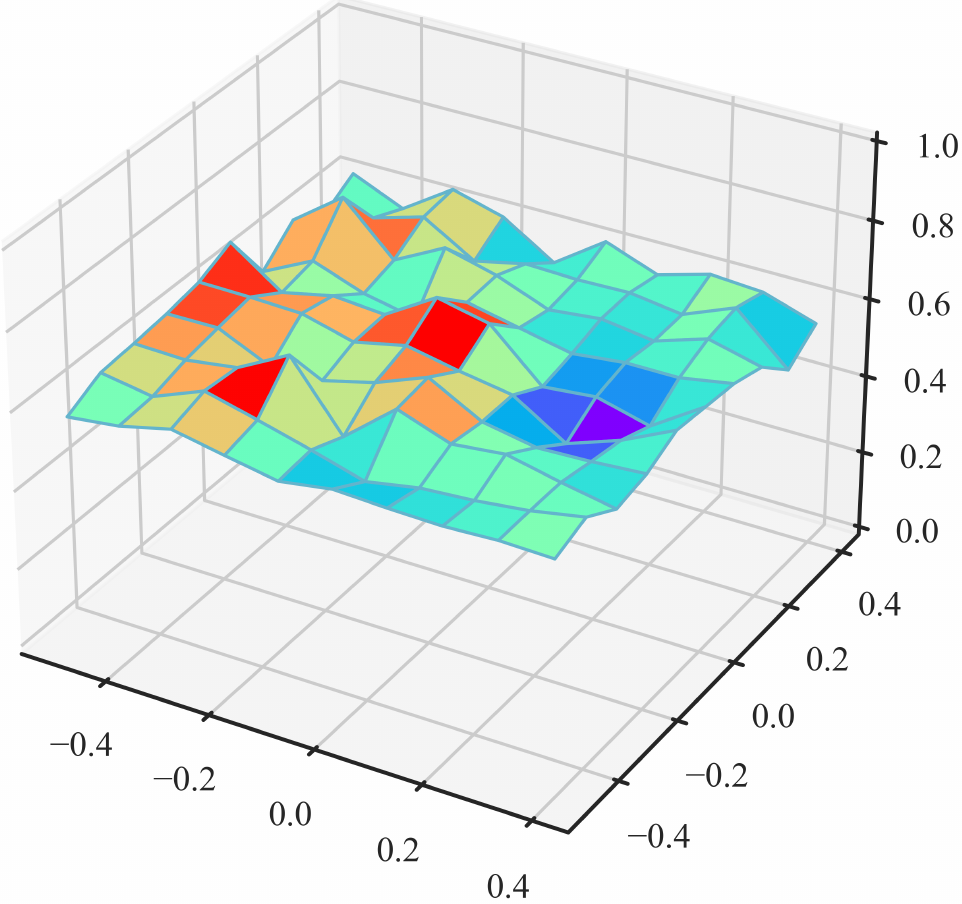}} 
	\caption{Output landscape on two-dimensional hyper-plane based on ResNet. specifically, one direction is the FGSM direction with a length of 1.0 pixels. Another direction is a random choice. Fig. \ref{fig:simple a} is the ResNet. Fig. \ref{fig:simple b} is the ResNet with CPRL. It can be found that the landscape is flat than original one. Regardless of the randomly selected perturbation direction or the perturbation direction of the adversarial attack, our landscape is flat, which empirically proves that CPRL is more robust.}
	\label{fig:landscape} 
\end{figure}

\begin{table*}[ht]
\centering
\caption{The SRCC and PLCC result of the existing IQA methods under 1/255 FGSM attack rates with clean training. }
\label{tab:fgsm}
\setlength{\tabcolsep}{3.5pt}
\begin{tabular}{c|ccc|ccc|ccc|ccc}
\toprule
\midrule
Datasets & \multicolumn{3}{c|}{VCL} & \multicolumn{3}{c|}{LIVE} & \multicolumn{3}{c|}{LIVEC} & \multicolumn{3}{c}{KONIQ} \\ \midrule
 & SROCC & PLCC & MSE & SROCC & PLCC & MSE & SROCC & PLCC & MSE & SROCC & PLCC & MSE \\  \midrule
DBCNN~\cite{zhang2018blind} & 0.1895 & 0.1677 & 0.0750 & 0.3216 & 0.3496 & 0.0658 & -0.3211 & -0.3502 & 0.1108 & -0.1383 & -0.1750 & 0.0420 \\
NIMA~\cite{talebi2018nima} & -0.3869 & -0.3838 & 0.1355 & 0.2278 & 0.2046 & 0.0740 & -0.2417 & -0.2656 & 0.1006 & 0.0805 & 0.0723 & 0.0290 \\
ResNet~\cite{he2016deep} & -0.0892 & -0.1045 & 0.0791 & -0.0335 & 0.0700 & 0.0985 & -0.1803 & -0.2102 & 0.0809 & -0.0831 & -0.1109 & 0.0372 \\
LWTA~\cite{panousis2021stochastic} & 0.2999 & 0.3386 & 0.1659 & 0.5840 & 0.2550 & 0.1739 & 0.2438 & 0.2434 & 0.0721 & 0.6060 & 0.5816 & 0.0328 \\
Denoise~\cite{xie2019feature} & -0.3285 & -0.3086 & 0.0971 & -0.1447 & -0.1082 & 0.1118 & -0.2677 & -0.2880 & 0.0826 & 0.1598 & 0.1522 & 0.0251 \\
SAT~\cite{xie2020smooth} & 0.3772 & 0.3409 & 0.0659 & 0.5100 & 0.5679 & 0.0528 & -0.0666 & -0.0555 & 0.0736 & -0.3177 & -0.3280 & 0.0581 \\
CPRL & \textbf{0.9041} & \textbf{0.8841} & \textbf{0.0134} & \textbf{0.9022} & \textbf{0.8785} & \textbf{0.0123} & \textbf{0.7675} & \textbf{0.8173} & \textbf{0.0144} & \textbf{0.8228} & \textbf{0.8562} & \textbf{0.0052} \\  
\bottomrule
\end{tabular} 
\label{tab:SOTA IQA}  
\end{table*}

\begin{table*}[ht]
\centering
\caption{The SRCC and PLCC result of the existing IQA methods under 1/255 PGD attack rates with clean training.}
\label{tab:pgd}
\setlength{\tabcolsep}{4pt}
\begin{tabular}{c|ccc|ccc|ccc|ccc}
\toprule
\midrule
Datasets & \multicolumn{3}{c|}{VCL} & \multicolumn{3}{c|}{LIVE} & \multicolumn{3}{c|}{LIVEC} & \multicolumn{3}{c}{KONIQ} \\ \midrule
 & SROCC & PLCC & MSE & SROCC & PLCC & MSE & SROCC & PLCC & MSE & SROCC & PLCC & MSE \\  \midrule
DBCNN & 0.1078 & 0.0815 & 0.0724 & 0.3881 & 0.3364 & 0.0582 & -0.3627 & -0.4214 & 0.1188 & 0.0628 & 0.0317 & 0.0447 \\
NIMA & -0.4032 & -0.3996 & 0.1310 & 0.1652 & 0.2090 & 0.0688 & -0.3841 & -0.4130 & 0.1154 & -0.0367 & -0.0578 & 0.0340 \\
ResNet & -0.0868 & -0.1226 & 0.0801 & 0.2346 & 0.3368 & 0.0744 & -0.2524 & -0.2648 & 0.0818 & -0.1245 & -0.1507 & 0.0371 \\
LWTA & 0.3539 & 0.3387 & 0.2252 & 0.5887 & 0.1171 & 0.1780 & 0.2567 & 0.2454 & 0.0659 & 0.6291 & 0.6213 & 0.0346 \\
Denoise & -0.2487 & -0.2345 & 0.0888 & 0.0401 & 0.1459 & 0.0833 & -0.2560 & -0.2618 & 0.0744 & 0.0676 & 0.0576 & 0.0265 \\
SAT & 0.2437 & 0.2060 & 0.0784 & 0.4961 & 0.5534 & 0.0554 & -0.2103 & -0.1984 & 0.0853 & -0.1507 & -0.1578 & 0.0369 \\
CPRL & \textbf{0.8694} & \textbf{0.8581} & \textbf{0.0176} & \textbf{0.8864} & \textbf{0.8530} & \textbf{0.0153} & \textbf{0.7295} & \textbf{0.7766} & \textbf{0.0173} & \textbf{0.8235} & \textbf{0.8542} & \textbf{0.0054} \\  
\midrule
\bottomrule
\end{tabular} 
\end{table*}

\section{Experiments}\label{sec:exp}


\subsection{Settings}
\paragraph{datasets}
We evaluate our method on four IQA datasets with different characteristics: LIVE and VCL, which contain artificially distorted images with a small size; and LIVEC and KONIQ, which contain naturally distorted images with a large size. KONIQ is the largest dataset with 10K images. \cref{tab:iqadata} summarizes the details of the datasets.

\begin{table}[ht]
\centering
\caption{Summary of IQA datasets.}
\fontsize{10pt}{12pt}\selectfont 
\scalebox{1.0}{
{
\begin{tabular}{cccc}
\hline 
\multirow{2}{*}{Databases} & \# of Dist. & \# of Dist.  & Distortions \tabularnewline
 & Images & Types & Type\tabularnewline
\hline 
LIVE \cite{sheikh2006liveiqa}& 799 & 5 & synthetic\tabularnewline  
VCL \cite{zaric2011vcl}& 575 & 4 & synthetic\tabularnewline
LIVEC \cite{ghadiyaram2015massive}& 1,162 & - & authentic\tabularnewline
KONIQ \cite{ghadiyaram2015massive}& 10,073 & - & authentic\tabularnewline
\hline 
\end{tabular}
}} 
\label{tab:iqadata}
\end{table}

\paragraph{Comparison methods}
We evaluate our proposed approach against three classic DNN-based IQA models and three adversarial robust models. As the first work to study attack and defense in the IQA domain, we select the simplest and most effective defence method: the LWTA method~\cite{panousis2021stochastic}, the denoising-based method~\cite{xie2019feature}, and SAT~\cite{xie2020smooth}.    
\paragraph{Attack methods}
In order to verify the effectiveness of the proposed method, we choose the most classic adversarial sample attack methods FGSM and PGD\cite{madry2017towards}, which are based on the gradient and are very representative.  

\paragraph{Metrics}
We use three widely used metrics to evaluate the performances of the models, i.e., the Pearson's Linear Correlation Coefficient (PLCC)~\cite{benesty2009pearson}, the Spearman's Rank Order Correlation Coefficient (SROCC)~\cite{zar2005spearman}, the Mean Square Error (MSE). 
\paragraph{Implementation} 
We split the data into training and testing sets at a 4:1 ratio randomly and saved the split-sequence to ensure the same division for all experiments. We also ensure that no source scene appears in both training and testing sets to avoid artificial inflation of the results. We resize the images in the training set to $320$ randomly and crop them to $320*320$ randomly. We resize and center crop the images in the test set. We assume that the cropped images have the same score as the original ones. We set the batch size to $16$ and use AdamW with a learning rate of $3e-5$ for optimization. We initialize $b$ to $0.4$.

\begin{table*}[ht]
\centering
\caption{The SRCC and PLCC result of the existing IQA methods without attack rates under clean training.}
\label{tab:clean}
\setlength{\tabcolsep}{4pt}
\begin{tabular}{c|ccc|ccc|ccc|ccc}
\toprule
\midrule
Datasets & \multicolumn{3}{c|}{VCL} & \multicolumn{3}{c|}{LIVE} & \multicolumn{3}{c|}{LIVEC} & \multicolumn{3}{c}{KONIQ} \\ \midrule
 & SROCC & PLCC & MSE & SROCC & PLCC & MSE & SROCC & PLCC & MSE & SROCC & PLCC & MSE \\ \midrule
DBCNN & 0.8967 & 0.8704 & 0.0277 & 0.9502 & 0.9274 & 0.0102 & 0.7157 & 0.7535 & 0.0191 & 0.8898 & 0.9116 & 0.0032 \\
NIMA & 0.9380 & 0.8890 & 0.0494 & 0.9396 & 0.9364 & 0.0105 & 0.7866 & 0.8292 & 0.0132 & 0.9014 & 0.9222 & 0.0029 \\
ResNet & 0.9234 & 0.9073 & 0.0115 & 0.8807 & 0.8212 & 0.0298 & 0.8341 & 0.8633 & 0.0121 & 0.8903 & 0.9123 & 0.0032 \\
LWTA & 0.2865 & 0.2907 & 0.2346 & 0.5259 & 0.1420 & 0.1788 & 0.2345 & 0.2521 & 0.0688 & 0.6384 & 0.6288 & 0.0375 \\
Denoise & 0.9317 & 0.9092 & 0.0158 & 0.8905 & 0.8238 & 0.0302 & 0.8191 & 0.8482 & 0.0131 & 0.8902 & 0.9155 & 0.0032 \\
SAT & 0.8021 & 0.7359 & 0.0315 & 0.8411 & 0.8146 & 0.0208 & 0.7005 & 0.7542 & 0.0193 & 0.8320 & 0.8680 & 0.0051 \\
CPRL & 0.9292 & 0.9222 & 0.0101 & 0.9266 & 0.9143 & 0.0104 & 0.8093 & 0.8620 & 0.0123 & 0.8567 & 0.8868 & 0.0042 \\  \midrule
\bottomrule
\end{tabular} 
 
\end{table*}

\subsection{Evaluations} 

\begin{table}[ht]
\centering  
\caption{Performance comparison under different b on the LIVE dataset.}\label{tab:b} 
\setlength{\tabcolsep}{2.5pt}
\begin{tabular}{c|ccc|ccc}
\toprule
\midrule
Datasets & \multicolumn{3}{c|}{clean} & \multicolumn{3}{c}{attack} \\ \midrule
b & SRCC & PLCC & MSE & SRCC & PLCC & MSE \\ \midrule
0.0 & 0.0100 & 0.0238 & 0.0534 & -0.0249 & -0.0663 & 0.0547 \\
0.1 & -0.2014 & -0.2147 & 0.0576 & -0.0296 & -0.0155 & 0.0543 \\
0.2 & 0.6798 & 0.6585 & 0.0400 & 0.5594 & 0.5303 & 0.0428 \\
0.3 & 0.8154 & 0.7618 & 0.0375 & 0.7933 & 0.7353 & 0.0359 \\
0.4 & 0.8560 & 0.8284 & 0.0350 & \textbf{0.8432} & \textbf{0.7898} & \textbf{0.0427} \\
0.5 & \textbf{0.8934} & \textbf{0.8646} & \textbf{0.0376} & -0.3538 & -0.3671 & 0.1450 \\  \midrule
\bottomrule
\end{tabular}
\end{table}


The existing IQA model and adversarial defence models have poor robustness, as shown by the results in \cref{tab:fgsm} and \cref{tab:pgd}. Their performance drops significantly with a small perturbation to the image under FGSM attack or PGD attack. The model's SRCC and PLCC values become negative, which means that the model gives inconsistent scores for slightly perturbed images. Our CPRL method enhances the robustness of the model and enables it to output more reliable scores for perturbed images. We can observe from \cref{tab:clean} that our prediction performance on natural samples is also superior, compared with other methods. Please see the Appendix for more experimental results.
 
\subsection{Ablation Studies}

In this section, we conduct experiments to study the efect on our approach of 1) Whether to use PNS loss 2) channel activation rate 3) attack rate
\paragraph{Impact of PNS risk}
To examine the effect of PNS risk loss, we contrasted the results with channel activation alone. As shown in table \cref{tab:PNS}, PNS achieves superior results. Both accuracy and robustness are enhanced.

\begin{table} 
\centering  
\caption{Performance comparison on the LIVE dataset.}\label{tab:PNS} 
\setlength{\tabcolsep}{2.5pt}
\begin{tabular}{c|ccc|ccc}
\toprule
\midrule
Datasets & \multicolumn{3}{c|}{clean} & \multicolumn{3}{c}{FGSM-1} \\
\midrule  
Model & SRCC & PLCC & MSE & SRCC & PLCC & MSE \\ \midrule
CA & 0.8743 & 0.8639 & 0.0163 & 0.8623 & 0.8478 & 0.0177 \\
CA+PNS & 0.9266 & 0.9143 & 0.0104 & 0.9022 & 0.8785 & 0.0123 \\ \midrule  
\bottomrule
\end{tabular}
\end{table}

\begin{figure}
  \centering
  \includegraphics[width=0.4\textwidth]{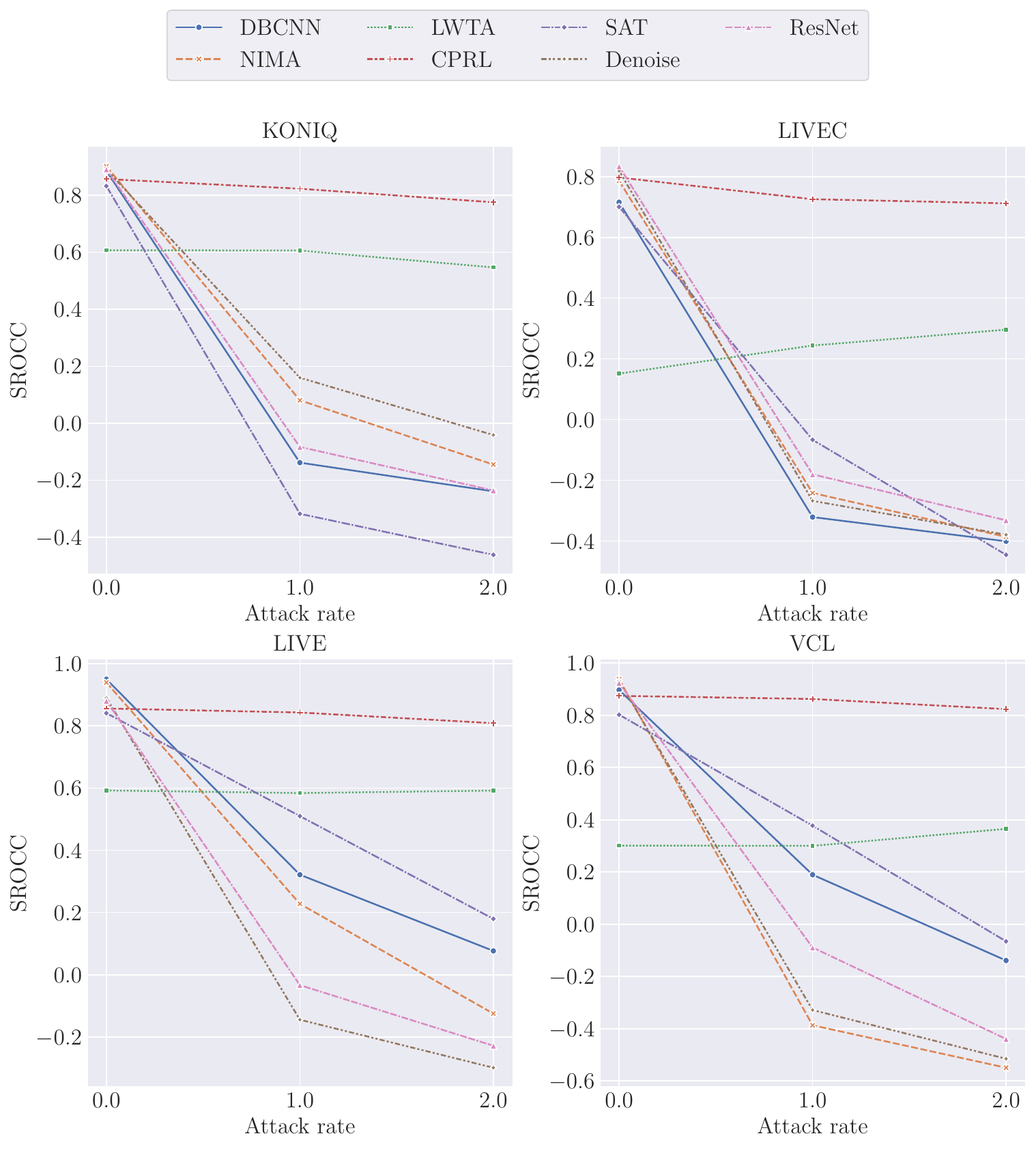}
  \caption{The SRCC result of the defence models under different FGSM attack rate with clean training.}\label{fig:attack rate}
\end{figure}

\paragraph{Impact of different attack rate}
We conducted experiments to analyze them under different testing (different attack rates).
As shown in \cref{fig:attack rate}, we set a uniform range of attack strengths to test the robustness of the proposed method. All models are all trained clean. Other methods suffer severe performance degradation under increasing attack intensity. By observing the results, it can be concluded that under different attack intensities, the proposed method can enhance the robustness of the model, and is superior to the comparison method.

\paragraph{Impact of channel activation rate} 
To investigate the effect of the activation ratio controlled by parameter $M $ on model performance, we conducted ablation studies on bias $b$. The experimental results are presented in table~\ref{tab:b}. We performed experiments on the LIVE dataset. The model is based on the ResNet50 network with channel activation and evaluated the robustness under different parameters. The experimental results reveal that the smaller the parameter b, the larger the activation ratio, which leads to better performance on clean samples, but lower robustness. We found that 0.4 is the optimal value for b.

\subsection{Case Study}
\begin{figure}
  \centering
  \includegraphics[width=0.4\textwidth]{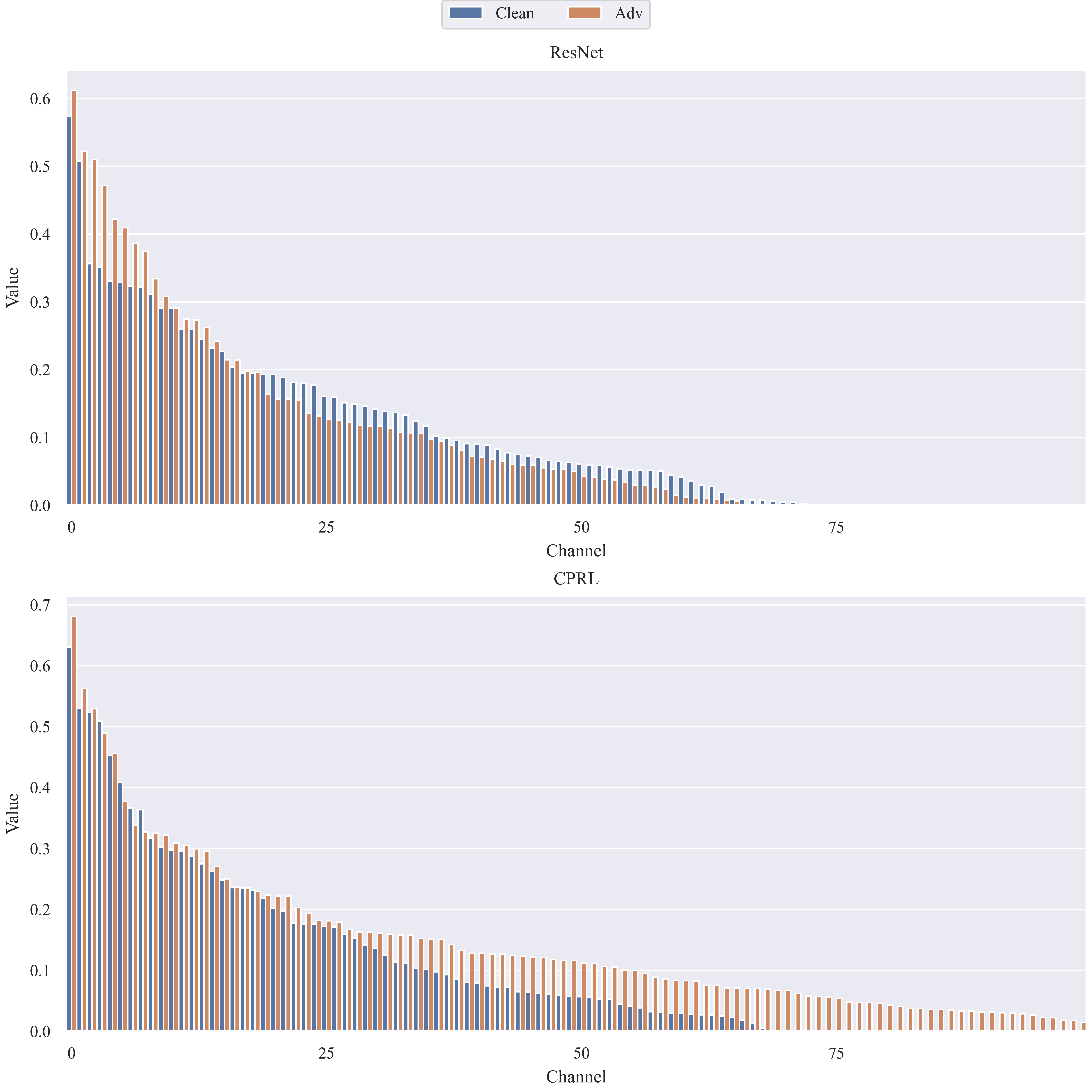}
  \caption{Channel activations value (y-axis) of intermediate layers of ResNet and CPRL models. In each figure, natural and adversarial test examples are shown respectively. The channels are arranged in descending order of magnitude. We found that CPRL values with larger magnitudes are more stable than the original ResNet. This is because we use PNS optimization to give larger values a higher perceptual correlation score, and non-perceptual perturbations have less impact on this part.}\label{fig:bar}
\end{figure}

\paragraph{Channel Activation value}  
\cref{fig:bar} illustrates the bar plot of channel-wise activations for natural and adversarial test samples generated by FGSM attacks. For the vanilla ResNet model (trained on natural samples), the adversarial perturbations induce substantial deviations in the activation patterns of individual channels, resulting in the propagation of adversarial noise from the input layer to the output layer of the network. The ResNet model enhanced by CPRL exhibits more robustness than the vanilla ResNet in regions with high activation values. The optimization process ensures that regions with higher activation values are more causally salient, thus mitigating the output variation.

\paragraph{landscape}  
\cref{fig:landscape} depicts the output landscape on a 2D hyperplane based on ResNet and CPRL. For vanilla ResNet, the landscape exhibits more variations, while CPRL shows more stability, and experiments confirm that CPRL is more robust.
 
\section{Conclusion}\label{sec:con} 
In this paper, we propose to build a trustworthy IQA model through causal-aware inspired representation learning (CPRL), which only requires inserting the channel activation module before the regular activation function and adopts PNS risk for optimization. Extensive experiments on popular quality assessment datasets verify that our method can effectively improve the adversarial robustness of quality assessment models. The current work opens the door to trustworthy robust IQA models.
\paragraph{Limitation and Future Work}  
Despite the strong performance of our model, it still suffers from some limitations. First, due to the more sophisticated optimization process, it requires additional optimization steps, which incur higher computational overhead. Moreover, achieving adversarial robustness through causal intervention is still an open challenge. The method of intervention through prediction in this paper may not be fully accurate, and there is still space for improvement. We intend to tackle these limitations and extend our approach in our future work.


{\small
\bibliographystyle{ieee_fullname}
\bibliography{main}
}
\end{document}